\documentclass{article}

\usepackage{cite}
\usepackage{amsmath,amssymb,amsfonts}
\usepackage{algorithmic}
\usepackage{textcomp}
\usepackage{adjustbox}
\usepackage{subcaption}
\usepackage{multirow}

\usepackage{bm}
\usepackage{makecell}

\usepackage{siunitx}

\usepackage{PRIMEarxiv}

\usepackage[utf8]{inputenc} 
\usepackage[T1]{fontenc}    
\usepackage{hyperref}       
\usepackage{url}            
\usepackage{booktabs}       
\usepackage{amsfonts}       
\usepackage{nicefrac}       
\usepackage{microtype}      
\usepackage{lipsum}
\usepackage{fancyhdr}       
\usepackage{graphicx}       
\graphicspath{{media/}}     

\title{Improving Failure Prediction in Aircraft Fastener Assembly Using Synthetic Data in Imbalanced Datasets
}

\author{
  Gustavo J. G. Lahr \\
  Hospital Israelita Albert Einstein \\
  São Paulo, Brazil\\
   \And
  Ricardo V. Godoy\\
  Engineering School of São Carlos\\
  University of São Paulo \\
  São Carlos, Brazil\\
  \And
  Thiago H. Segreto\\
  Engineering School of São Carlos\\
  University of São Paulo \\
  São Carlos, Brazil\\
  \And
  José O. Savazzi\\
  Federal University of São Carlos\\
  São Carlos, Brazil\\
  \And
  Arash Ajoudani\\
  Italian Institute of Technology \\
  Genoa, Italy\\
  \And
  Thiago Boaventura\\
  Engineering School of São Carlos\\
  University of São Paulo \\
  São Carlos, Brazil\\
  \And
  Glauco A. P. Caurin\\
  Engineering School of São Carlos\\
  University of São Paulo \\
  São Carlos, Brazil\\
}

\begin{document}
\maketitle

\begin{abstract}
Automating aircraft manufacturing still relies heavily on human labor due to the complexity of the assembly processes and customization requirements. One key challenge is achieving precise positioning, especially for large aircraft structures, where errors can lead to substantial maintenance costs or part rejection. Existing solutions often require costly hardware or lack flexibility. Used in aircraft by the thousands, threaded fasteners, e.g., screws, bolts, and collars, are traditionally executed by fixed-base robots and usually have problems in being deployed in the mentioned manufacturing sites. 
This paper emphasizes the importance of error detection and classification for efficient and safe assembly of threaded fasteners, especially aeronautical collars. Safe assembly of threaded fasteners is paramount since acquiring sufficient data for training deep learning models poses challenges due to the rarity of failure cases and imbalanced datasets. The paper addresses this by proposing techniques like class weighting and data augmentation, specifically tailored for temporal series data, to improve classification performance. 
Furthermore, the paper introduces a novel problem-modeling approach, emphasizing metrics relevant to collar assembly rather than solely focusing on accuracy. This tailored approach enhances the models' capability to handle the challenges of threaded fastener assembly effectively.
\end{abstract}


\section{Introduction}
\label{sec:introduction}
Aircraft manufacturing continues to rely heavily on human labor despite notable technological advancements \cite{Gonzaga2019}. Although some companies have made automation attempts \cite{Airbus2019assembly}, the outcomes have been mixed \cite{Gates2019}. This predicament arises due to the semi-unstructured nature of the manufacturing environments and the high degree of customization required for various parts \cite{Kheddar2019humanoidsaircraft}. Consider, for instance, the Boeing 747-8, which encompasses nearly 6 million components \cite{Boeing2013nparts}. The stringent quality standards further diminish the room for errors. Consequently, developing flexible and intelligent systems capable of detecting and promptly reacting to errors becomes crucial for achieving successful automation across all stages of assembly.

A critical task within this context is the assembly of threaded fasteners, including screws, bolts, and collars\footnote{Collars are a specific type of nut utilized in aircraft manufacturing, designed to shed part of their structure to ensure the necessary torque and reduce weight, rendering them as challenging to remove as rivets~\cite{Lisi-aerospace2021}.}. In a typical automated production line, fixed-base robots execute the screwing task. However, it demands precise positioning of both the robot and the component alongside an interaction controller \cite{Kuka2021screw1, Kuka2021screw2, STP-CONCEPT2021screw3, Duepi2021screw4, Denso2021screw5}. Given the substantial structure size, this approach proves impractical in aircraft manufacturing due to the requirement for in-place assembly. Achieving precise positioning remains a significant challenge, increasing production error risk. Existing solutions often rely on costly hardware \cite{Drouot2018} or yield systems with limited flexibility \cite{Mosqueira2012}.

In assembling large parts with numerous fasteners, even a single faulty assembly can result in substantial maintenance costs or part rejection due to safety concerns \cite{FAA2012, Hardy2019}. Such failures pose particular challenges in the case of permanent joints like collars, as robots struggle to react effectively due to difficulties in directly obtaining the current state. For instance, occlusion and hidden features render cameras ineffective in distinguishing crucial information \cite{Mason2018}. Therefore, a successful approach must leverage readily available data, such as kinematic data from the robot's end-effector and dynamic data from the force-torque sensors mounted on the robot's wrist.

Accurate error detection and classification can enable the system to recover from faulty situations, leading to a more efficient and safer assembly. The techniques associated with error detection fall under the umbrella of failure detection and isolation (FDI), which can be categorized into two main branches: model-based and data-driven. Model-based approaches, while efficient, require substantial effort and expertise to comprehend and analyze force/torque signatures \cite{Mason2010forcesignature, McIntyre2005, Arrichiello2015}. On the other hand, data-driven classification methods allow for directly utilizing raw data, simplifying implementation efforts, and expediting real-world deployment \cite{Diryag2014, Van2013, Ferhat2021}. Data-driven learning approaches have demonstrated their viability in classifying and predicting the outcomes of screwing tasks, enabling potential failures to be anticipated and circumvented during task execution \cite{Moreira2018, Cheng2019}.

Deep learning models excel in data-driven classification problems and perform better when provided with large datasets. However, acquiring sufficient data for training these models poses challenges as it is time-consuming, and the occurrence of failure cases varies across different phases of the assembly process, resulting in low rates, as low as 0.4\% in certain instances \cite{Aronson2017, Cheng2018}. Moreover, the datasets for screwing tasks are often limited in size and imbalanced between classes \cite{Aronson2017}. The small dataset size restricts the complexity of models that can be effectively employed, and training can be biased due to uneven occurrences across different classes \cite{Barella2021andrePonce}. Consequently, addressing the imbalanced data issue becomes imperative when assembling threaded fasteners.

\begin{figure}
    \centering
    \includegraphics[width=.75\linewidth]{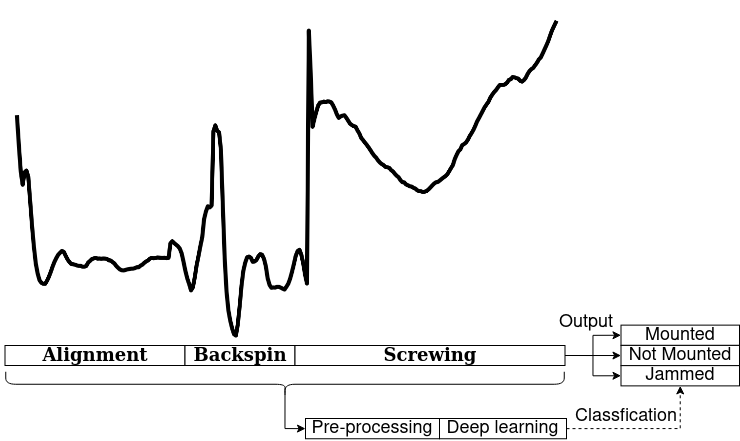}
    \caption{The collected data from the three phases (Alignment, Backspin, and Screwing) are pre-processed and used as input for several deep-learning models. The outputs are split between the three possible outcomes: Mounted, Not mounted, and Jammed.}
    \label{fig:task_description}
\end{figure}

This paper builds upon our previous work \cite{Moreira2018} and presents two significant contributions that advance the state-of-the-art in failure detection and isolation (FDI) during the process of screwing collars. The first contribution is enhancing multi-class time-series classification for small screwing datasets using deep learning models. We employ raw data inputs with standard pre-processing and compare two techniques for addressing the challenges of imbalanced datasets: class weighting and data augmentation specifically designed for temporal series. Additionally, we evaluate the performance of the models with synthetic augmentation and demonstrate that while artificial data improves classification metrics, generating an order of magnitude of synthetic data does not yield further improvements in failure detection.

The second contribution of this paper introduces a novel problem-modeling approach for screwing tasks. Rather than selecting deep learning models based solely on accuracy, which can mask problems arising from imbalanced data, we propose optimizing the models based on metrics specific to the collar assembly process. This tailored approach enhances the models' capability to effectively handle the challenges of assembling threaded fasteners.

To further improve the effectiveness of our approach, we implemented several minor enhancements. Firstly, we employ a hyperparameter optimization pipeline for temporal series, enabling us to select the optimal model architecture. In conjunction with a k-fold split, this approach minimizes the model's sensitivity to the dataset, ensuring robust performance. Additionally, we incorporate the state-of-the-art Vision Transformer model \cite{dosovitskiy2021image, Godoy2022}, comparing its performance with other baseline models, such as fully connected, recurrent, convolutional, and other transformer-based networks prevalent in the literature. Lastly, we investigate the impact of including kinematic information in addition to force/torque data, allowing us to evaluate the potential benefits of integrating multiple data modalities.

This paper is organized as follows: Sec.~\ref{sec:related-works} presents an overview of the state-of-the-art regarding model-based and data-driven techniques for FDI classification, with a special focus on screwing classification; Sec.~\ref{sec:methods} introduces the problem modeling used in this paper, together with the dataset description and the pipeline used to train and evaluate the models; finally, Sec.~\ref{sec:results} shows the results obtained and discusses the best practices for collar screwing detection, with conclusions and future directions in Sec.~\ref{sec:conclusions}.

\section{Related works}\label{sec:related-works}

Several works in the literature have proposed model-based approaches for assembling threaded fasteners. Notably, Seneviratne et al. \cite{Seneviratne2001} and Wiedmann et al. \cite{Wiedmann2006spatialKinAnalysis} have developed models for self-tapping screws and machine bolts, respectively. Wiedmann \cite{Wiedmann2006threadStarting} specifically focused on modeling the thread start of bolts to gain insights into the occurrence of cross-threading problems, which are known challenges in threading tasks \cite{Mason2018}. Nicolson \cite{Nicolson1991, Nicolson1993} proposed a dynamic model for the robotic assembly of threaded fasteners, investigating a compliant interaction controller for threading tasks. However, despite the benefits of these models in designing assembly strategies, their practical deployment in production systems is hindered by the impracticality of instrumenting the necessary measurements directly and the reliance on indirect measurement techniques. Consequently, data-driven approaches have gained prominence, enabling the learning of nonlinear relationships between input data and the assembly state.

Regarding data-driven approaches, early studies focused on binary classification between successful and unsuccessful screwing attempts using artificial neural networks. Lara et al. \cite{Lara2000} utilized tightening torque and insertion depth to manually assemble self-tapping threaded fasteners. In contrast, Althoefer et al. \cite{Althoefer2005} expanded on this work by incorporating tightening torque and rotation angle as inputs. Ruusunen et al. \cite{Ruusunen2003} employed fuzzy reasoning to differentiate between successful and unsuccessful screwing states using torque as the input for a robotic screwing task.

However, binary classification alone is inadequate for automation, as different types of faults (e.g., positioning error, cross-threading, missing bolt/nut) require distinct recovery processes. Consequently, multiclass classification approaches have been investigated. Althoefer et al. \cite{Althoefer2008} employed artificial neural networks for multiclass faulty case classification in manual self-tapping screw assembly. Matsuno et al. \cite{Matsuno2013} analyzed robotic screw-driving tasks using a support vector machine (SVM) as the classifier. Aronson et al. \cite{Aronson2017} collected a relatively large dataset for a robotic screwdriver of miniature screws, extracting 85 different features based on time and comparing logistic regression with a Graph of Temporal Constraint Decision Forest. Building on Aronson et al.'s work, Cheng et al. \cite{Cheng2018} implemented a feature selector to eliminate inputs that had limited contribution to the prediction, resulting in 18 features obtained from motor current, z-axis force, and rotation angle. Decision tree models outperformed others regarding long short-term memory (LSTM), SVM, and linear regression regarding performance and interpretability. Teixeira et al. \cite{Teixeira2022} employed Principal Component Analysis to create the necessary features for their models.

Recognizing that errors during automated threading tasks often surpass those encountered during supervised learning model training, some researchers have explored unsupervised learning approaches. Ferhat et al. \cite{Ferhat2021} trained Gaussian mixture models to cluster faulty cases, and when all classes rejected a case, it was considered a new case. Cheng et al. \cite{Cheng2019} proposed a hybrid approach where previously known classes were predicted, where new generic error classes were created and grouped using hidden Markov models when the accuracy fell below a certain threshold. 

Due to the lack of data, the works in the literature usually extract temporal features to use as input for the models, enabling lighter models to be used~\cite{Cheng2019}. By doing so, the authors assume knowledge about the assembly process, but deployment in the assembly lines may lead to new types of errors not predicted in the laboratory. Using raw data is an approach that does not make any assumptions about the data, but the small datasets pose a challenge. In our previous work \cite{Moreira2018}, we compared three methods—multilayer perceptron (MLP), SVM, and convolutional neural network (CNN)—for the FDI task using raw data as input. That study focused primarily on model accuracy, and a comparison was performed using confusion matrices. It is important to note two key observations from our previous work: none of the three classifiers was capable of accurately predicting jammed situations, and all classifiers exhibited a false positive tendency for the mounted class, with the majority of jammed cases (approximately 80\% for all models) being incorrectly classified as mounted. Furthermore, given the imbalanced nature of the dataset, using a global performance metric such as accuracy may mask certain classification issues.

Importantly, the existing works in the literature have not addressed the impact of data augmentation specifically for screwing tasks, nor have they adequately considered the evaluation of performance metrics tailored to the collar assembly process. Pastor et al. \cite{Pastor2021threadQualityImbalanced} explored techniques for handling imbalanced datasets in the binary classification of automated tapping with CNC, employing feature extraction approaches such as the area under the torque. In light of these research gaps, this paper aims to contribute novel methodologies and insights to the field, specifically focusing on applying data augmentation techniques and utilizing metrics tailored to the collar assembly process.

\section{Materials and Methods} \label{sec:methods}

This section outlines the methodology used. Due to the limited availability of failure cases and the inherent imbalance in real-world datasets, we adopt data augmentation and cost-sensitive training techniques. Using raw sensory data from a compliant robotic setup, we evaluate multiple deep learning models within a hyperparameter optimization framework, including convolutional, recurrent, and transformer-based architectures. The goal is to develop a task-oriented prediction system that enhances classification performance without feature engineering, particularly for critical failure modes.

\subsection{Dataset and Problem Modeling}

The dataset consists of 479 samples assembled with a Kuka KR16 robotic manipulator with an admittance controller obtained via a force-torque sensor mounted between the gripper and the robot's flange. It has 479 samples split into 306 (63.9\%) mounted cases, 112 (23.4\%) not mounted, and 61 (12.7\%) jammed. The dataset contains all time steps of forces and torques in all directions, plus displacements and rotation angles. We synchronize all samples by the initial contact between the collar and screw until the final screwing turn. More details are given in~\cite{Lahr2023dataset}. 

\begin{figure}
    \centering
    \includegraphics[width=0.7\linewidth]{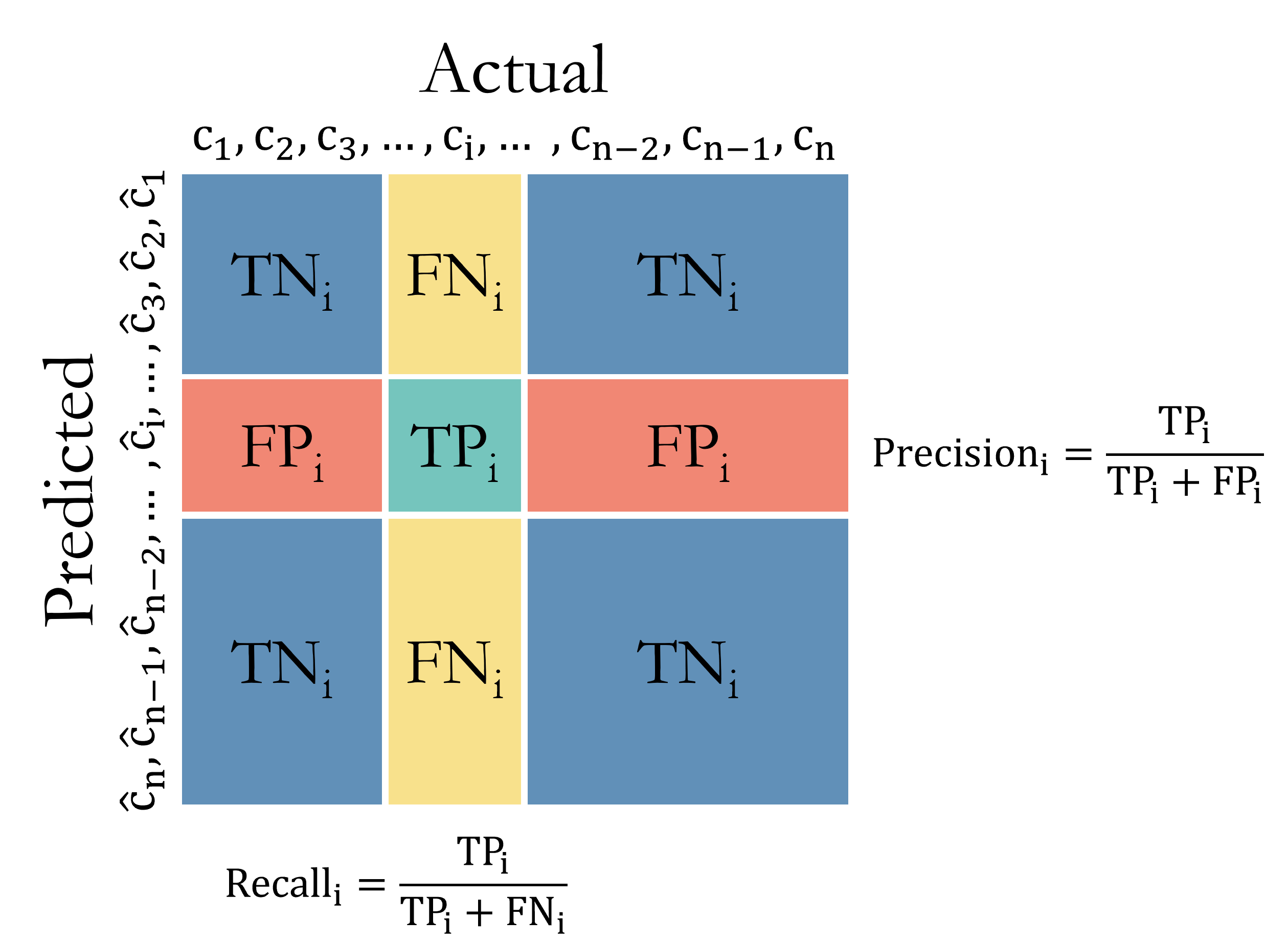}
    \caption{TP are used to account for accuracy}
    \label{fig:multiclass-confusion-matrix}
\end{figure}

Two additional analyses were conducted to improve the models. First, we changed the metric of interest from global accuracy to metrics related to a specific class commonly used for imbalanced datasets~\cite{Mortaz2020, Barella2021andrePonce}. For a problem with $n$ classes, there are four possible outcomes for the prediction of the $i$-th class: true positive (TP), predicted class is the actual value; true negative (TN), prediction as false and it is; false positive (FP), a true prediction that is another class; and false negative (FN), a false prediction that was supposed to be true. While accuracy is the ratio between all correct outputs by all the inputs ($ACC=(TP+TN)/(TP+FP+TN+FN)$), Precision and Recall are metrics associated with the $i$-th class $c_i$: precision expresses the proportion of true positives that are positives, and recall gives the ratio of positive instances that are correctly detected by the classifier. Figure~\ref{fig:multiclass-confusion-matrix} shows a graphical depiction of the metrics calculation with its confusion matrix.  

As the dataset is small, we propose a task-oriented classification that will leverage the most crucial cases. Therefore, two goals are used to define the metrics:
\begin{itemize}
    \item[1] \textit{Improve the Precision of the \textbf{mounted} class}: if the model predicts that it will mount but gets jammed instead (FP), it is worse than predicting that it will not mount or jam (FN) but would mount.
    \item[2] \textit{Improve Recall regarding the \textbf{jammed} class}: if the prediction is not jammed and instead gets jammed (FN), it is also problematic.
\end{itemize}

Moreira et al.~\cite{Moreira2018} have all models resulting in null recall, and the precision of MLP, SVM, and CNN is $80.5\%$, $80.2\%$, and $80.5\%$, respectively. These numbers highlight the need to improve the models' performance because, in real-world scenarios, such as aircraft manufacturing, where hundreds of thousands of threaded fasteners, even a small percentage of failures, can result in significant issues. To address this, this work prioritized maximizing precision metrics, particularly for the majority class, as it is the most likely outcome from the model. However, the final model selection should also consider its performance in terms of recall, specifically for identifying jammed fasteners, which are of particular concern. The F$_1$ metric was not used since precision and recall are considered from different classes.

Second, we added recurrent networks, which were not evaluated in~\cite{Moreira2018}. Recurrent networks were developed to deal with sequential data, i.e., series correlating with previous time steps. Another essential feature is that they can receive input of variable lengths, which is especially interesting for online applications. Since we are using raw data as input, we dropped SVM and added the long short-term memory (LSTM), Transformer~\cite{Vaswani2017}, and Temporal Multi-Channel Vision Transformers (ViT)~\cite{Godoy2022}. 

\begin{figure*}
    \centering
    \includegraphics[width=\linewidth]{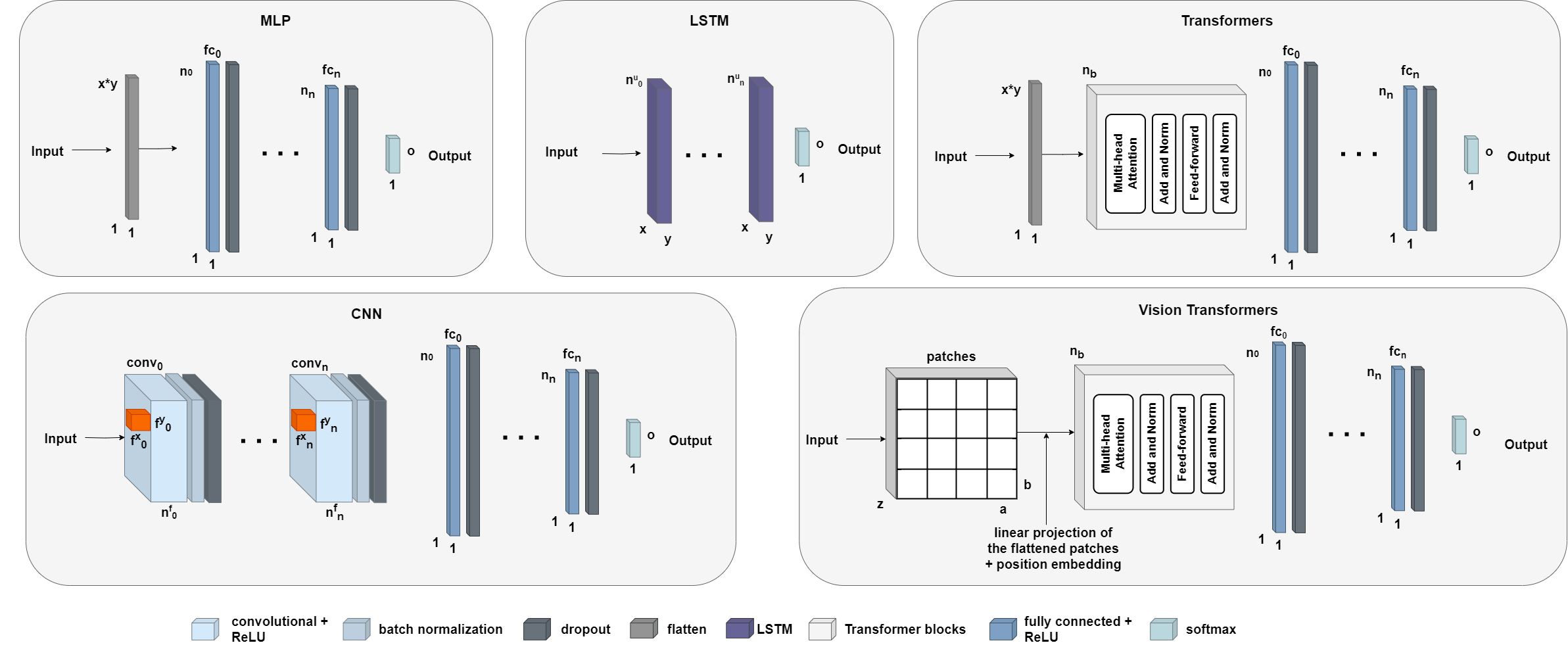}
    \caption{General architecture of the deep learning models evaluated in this work. MLP processes flattened time-series input through dense layers; CNN applies 1D convolutions to extract local temporal patterns; LSTM captures sequential dependencies across time steps; Transformer leverages self-attention for long-range dependencies; and ViT uses a temporal patch-based approach with convolutional token embedding and Transformer blocks.}

    \label{fig:models}
\end{figure*}

Figure~\ref{fig:models} shows the general architecture for each model. All models process raw time-series input representing force, torque, and kinematic signals from the robotic screwing task. The MLP architecture consists of fully connected layers applied directly to flattened input sequences. The CNN model extracts temporal features through stacked 1D convolutional and pooling layers. The LSTM architecture leverages recurrent units to capture sequential dependencies across time steps. The Transformer model employs self-attention mechanisms for modeling long-range dependencies. At the same time, the ViT architecture divides the input into temporal patches and processes them through multi-head attention layers, enabling rich cross-channel and temporal representation learning. Each model concludes with dense layers for multi-class classification.

Table~\ref{tab:hyperparams} details each evaluated model's search ranges used during hyperparameter optimization. The parameters include:

\begin{itemize}
    \item $n\ell_{fc}$: Number of fully connected (dense) layers.

    \item $nn_{fc}$: Number of neurons per fully connected layer, ranging from 1 to 2048.

    \item $Drop_{fc}(p)$: Dropout probability applied to fully connected layers to prevent overfitting, with values between 0 and 0.5.

    \item $\ell_{2_{fc}}$: $L_2$ regularization coefficient applied to the fully connected layers to control weight magnitude.

    \item $n\ell_{dn}$: Number of convolutional layers (for CNN, Transformer, and ViT models).

    \item $k_{dn}$: Kernel size of the convolutional layers.

    \item $Pool_{dn}$: Pooling size used in max or average pooling operations.

    \item $nn_{Tr}$: Size of the Transformer’s latent vector or embedding dimension, explored as powers of 2 from $2^4$ to $2^8$.
\end{itemize}

Note that for models that do not include certain components (e.g., convolutional layers in MLP), the corresponding entries are marked with a dash (–). This configuration allows each model to be flexibly and efficiently tuned to best fit the task of classifying outcomes from time-series sensor data.

\begin{table*}[] \caption{Hyperparameter search ranges used during model optimization. Each model is tuned independently using Optuna, with ranges defined for the number of layers, neurons per layer, dropout probability, regularization, convolutional settings, and Transformer dimensions. Notation follows standard conventions where applicable.} \label{tab:hyperparams}
    \centering
\begin{adjustbox}{width=\textwidth}
\begin{tabular}{lccccccccc}
Model \textbackslash HyP & $n\ell_{fc}$ & $nn_{fc}$   & $Drop_{fc}(p)$ & $\ell_{2_{fc}}$            & $n\ell_{dn}$ & $k_{dn}$      & $Pool_{dn}$ & $nn_{Tr}$ \\ \hline
MLP                      & $[1;10]$     & $[1; 2048]$ & $[0;0.5)$      & $[1\mathrm{e}{-3};0.1]$    & $-$          & $-$           & $-$       & $-$ \\ \hline
CNN                      & $[1;6]$      & $[1; 2048]$ & $[0;0.5)$      & $[1\mathrm{e}{-3};0.1]$    & $[1;8]$      & $\{1,3,5\}$   & $\{1,2\}$ & $-$ \\ \hline
LSTM                     & $\{1\}$      & $\{1\}$     & $[0;0.5)$      & $-$                        & $[1;5]$      & $-$           & $-$       & $-$ \\ \hline
Transf.                  & $\{1\}$      & $\{64\}$    & $[0;0.5)$      & $-$                        & $[1;8]$      & $\{8, 16\}$   & $-$       & $\{2^n, n\in[4,8]\}$ \\ \hline
ViT                      & 	$[1;4]$     & $\{64\}$    & $[0;0.5)$      & $-$                        & $\{5\}$      & $\{2, 3\}$    & $\{2, 3\}$  & $\{2^n, n\in[4,8]\}$ \\ \hline
\end{tabular}
\end{adjustbox}
\end{table*}

\subsection{Training and validation pipeline}

Two analyzes were conducted: first, we try to predict the outcome of the task with only the forces and torques; second, the classification of the current state by adding the screwing angle. 
We use a compliant assembly, i.e., the robot has an impedance controller constantly correcting trajectory based on high forces. This differs from other literature works that rely on position control and only implement force monitoring~\cite{Aronson2017, Cheng2019, Ferhat2021}. This approach is more general when considering other screwing tasks that a robot could implement without a screwdriver, such as assembling bottle and jar caps, power screws, wheel studs, and others.

\begin{figure}
    \centering
    \includegraphics[width=.98\linewidth]{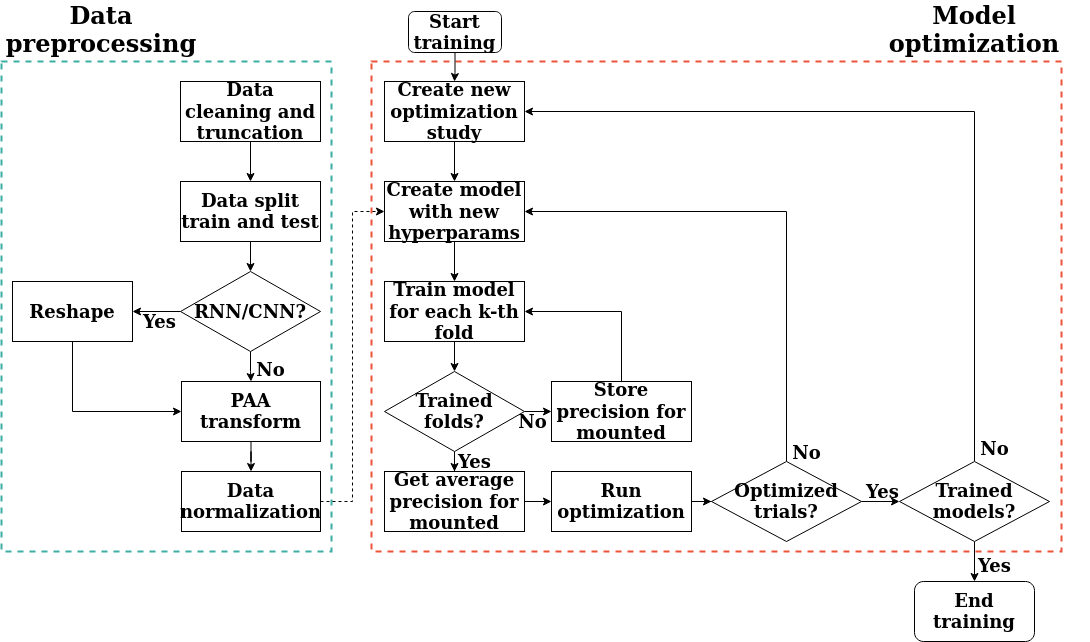}
    \caption{Hyperparameter optimization pipeline to obtain the tuned models. The test data is split from the train right after \textit{Data Normalization}. The \textit{model optimization} block runs for each model in different processes.}
    \label{fig:pipeline-ml}
\end{figure}

Since the dataset is small, the models are susceptible to a high variance of the input data. We deal with this problem using a hyperparameter optimization framework, named Optuna~\cite{optuna_2019}. For each step in the optimization, we train in a data split of 10-fold and get the mean of the mounted state precision. Figure~\ref{fig:pipeline-ml} shows an overview of the whole pipeline. The \textit{data preprocessing} phase has outlier cleaning and time series truncation for a fixed number of time steps split between train (80\%) and test (20\%) data, with an approximation of the time series using Piecewise Aggregate Approximation (PAA) transform that works as a filter and data normalization~\cite{ChaoHong2017paa}. Meanwhile, \textit{model optimization} runs for all models with 100 trials for optimized hyperparameters.  

\subsection{Imbalanced dataset}

We also studied the impact of synthetic data on the classification of the mentioned models, a valuable technique to help deal with imbalanced datasets, a common practice in other fields~\cite{Shorten2019dataAugComputerVision, Liu2020dataAugNLP}. Imbalanced datasets pose a challenge in machine learning applications, with special attention to time series analysis, where the time dimension adds an extra layer of complexity to the problem. Unlike static datasets, time series data exhibit inherent temporal dependencies that render conventional solutions ineffective. For instance, datasets in robotics are often scarce and highly imbalanced since it is expensive to obtain failure cases. Unlike in traditional machine learning contexts, under-sampling is often impractical due to the already limited size of these datasets. This combination of imbalanced class distribution and small-sized datasets is a focus of this paper and is addressed in the following sections.

\subsubsection{Data augmentation}

Introduced by~\cite{Chawla2002smote}, SMOTE was proposed for oversampling the minority class by generating synthetic data by selecting a random instance from the minority class and then the k nearest minority class neighbors. One of the neighbors is chosen randomly, and a vector related to the Euclidean distance between the element and its neighbors is calculated. A new artificial sample is produced by multiplying this vector by a random real number from the 0 to 1 range. This procedure is repeated until the desired number of synthetic samples is reached, extending the decision area of the minority class. 

Thus, three datasets were used to train new models: the original, as published in~\cite{Moreira2018}; a balanced dataset, where jammed and not mounted classes were synthesized until they had the same sample amount as the mounted class, i.e., 243 samples for each class used for training; and a third dataset, named synthetic, which we took the balanced dataset then we created four times more data, leading to a total of 972 samples for each class, 2,916 in total. Testing remained the same as the experimental 96 samples.

\subsubsection{Class weights}

To compare different techniques, we also added the evaluation of class weights during model training. Class weights can be used to address imbalanced datasets in machine learning. This approach involves assigning different weights to classes based on their prevalence in the training dataset. By doing so, the model focuses on the minority class, thereby enhancing its performance on this class without altering the original dataset or generating synthetic data~\cite{johnson2019survey}. Our study employed the direct method approach to incorporate class weights into the training process~\cite{ling2008cost}. Direct methods inherently possess cost-sensitive capabilities, achieved by modifying the learner's underlying algorithm to consider the costs associated with class weights during learning. Consequently, the optimization objective shifts from minimizing total error to minimizing total cost. We utilized a weight vector based on the sample counts of each class.

\section{Results and Discussion}\label{sec:results}

\begin{figure}
    \centering
    \subfloat[\centering]{
        \includegraphics[width=.975\linewidth]{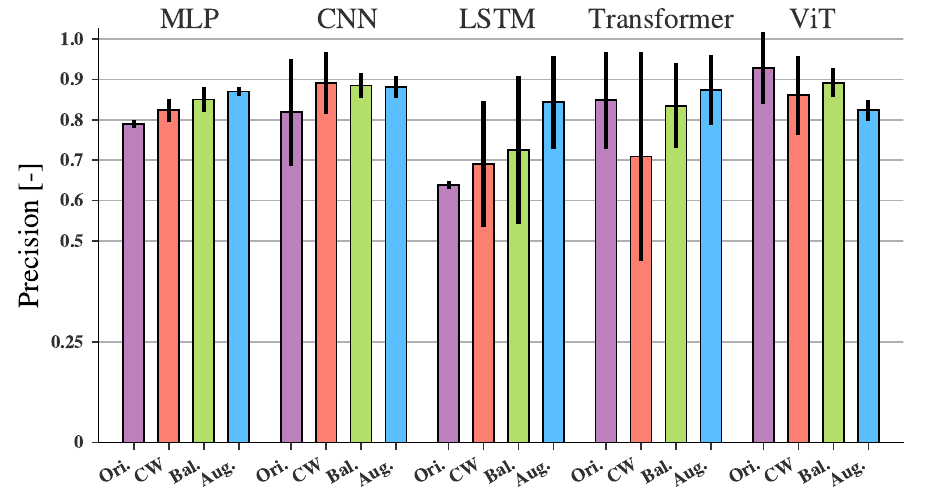}}\\
    \subfloat[\centering]{
        \includegraphics[width=.975\linewidth]{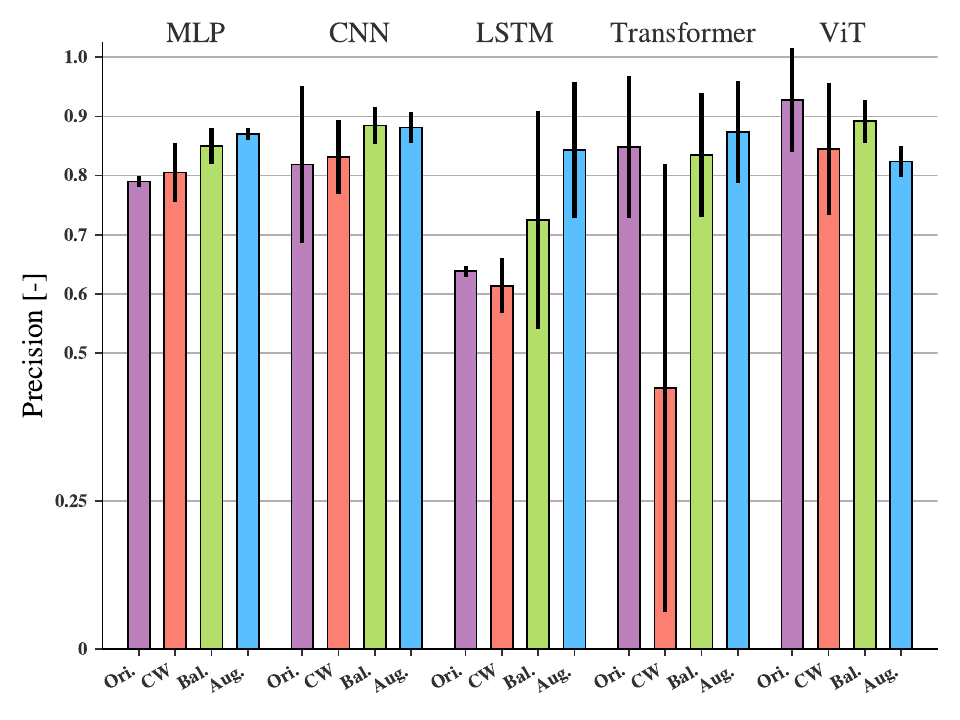}}
    \caption{Mean precision and standard error for the mounted class obtained during hyperparameter optimization of each model for (a) without rotation and (b) with rotation data. They are evaluated across three dataset variants: original, balanced, and synthetic, as well as a class weighing technique. Results reflect 10-fold cross-validation averaged over 100 optimization trials. }.
    \label{fig:optimized-ml-models}
\end{figure}

This section presents the performance evaluation of the trained models for predicting the outcome of threaded fastener assembly. We analyze the results based on task-specific metrics (considering the impact of rotational data), model complexity, and data imbalance strategies. All results reported are the mean±standard deviation across 10-fold cross-validation. Table~\ref{tab:precision-recall} shows the hyperparameter optimization results, and Figure~\ref{fig:optimized-ml-models} displays the bar plots for the mounted class precision. Adding more data improved the models by increasing the average mounted precision or lowering the standard deviation.

\begin{table*}[ht]
    \centering
    \caption{Mounted Precision (Pr) and Jammed Recall (Re) for each model, data treatment, and rotation condition. Values are means $\pm$ standard deviation.}
    \label{tab:precision-recall}
    \sisetup{table-align-uncertainty=true, separate-uncertainty=true}
    \resizebox{\textwidth}{!}{
    \begin{tabular}{@{} l *{4}{S[table-format=1.2(2)] S[table-format=1.2(2)]} @{}}
        \toprule
        & \multicolumn{2}{c}{\textbf{Original}} & \multicolumn{2}{c}{\textbf{Original + CW}} & \multicolumn{2}{c}{\textbf{Balanced}} & \multicolumn{2}{c}{\textbf{Synthetic}} \\
        \cmidrule(lr){2-3} \cmidrule(lr){4-5} \cmidrule(lr){6-7} \cmidrule(lr){8-9}
        \textbf{Model} & {Pr} & {Re} & {Pr} & {Re} & {Pr} & {Re} & {Pr} & {Re} \\
        \midrule
        \multicolumn{9}{c}{\textit{Without rotation}} \\
        \midrule
        MLP         & 0.79 \pm 0.01 & 0.00 \pm 0.00 & 0.86 \pm 0.11 & 0.81 \pm 0.30 & 0.85 \pm 0.03 & 0.00 \pm 0.00 & 0.87 \pm 0.01 & 0.00 \pm 0.00 \\
        CNN         & 0.87 \pm 0.11 & 0.25 \pm 0.08 & 0.74 \pm 0.07 & 0.20 \pm 0.21 & 0.90 \pm 0.03 & 0.25 \pm 0.04 & 0.89 \pm 0.02 & 0.42 \pm 0.06 \\
        LSTM        & 0.71 \pm 0.15 & 0.00 \pm 0.00 & 0.63 \pm 0.10 & 0.37 \pm 0.26 & 0.70 \pm 0.11 & 0.17 \pm 0.15 & 0.76 \pm 0.12 & 0.08 \pm 0.15 \\
        Transformer & 0.75 \pm 0.15 & 0.00 \pm 0.00 & 0.40 \pm 0.38 & 0.34 \pm 0.43 & 0.87 \pm 0.14 & 0.42 \pm 0.12 & 0.86 \pm 0.08 & 0.75 \pm 0.09 \\
        ViT         & 0.95 \pm 0.10 & 0.00 \pm 0.00 & 0.79 \pm 0.28 & 0.02 \pm 0.03 & 0.87 \pm 0.03 & 0.33 \pm 0.10 & 0.83 \pm 0.01 & 0.17 \pm 0.08 \\
        \midrule
        \multicolumn{9}{c}{\textit{With rotation}} \\
        \midrule
        MLP         & 0.79 \pm 0.01 & 0.00 \pm 0.00 & 0.81 \pm 0.05 & 0.57 \pm 0.26 & 0.85 \pm 0.03 & 0.00 \pm 0.00 & 0.87 \pm 0.01 & 0.00 \pm 0.00 \\
        CNN         & 0.82 \pm 0.13 & 0.00 \pm 0.00 & 0.83 \pm 0.06 & 0.48 \pm 0.29 & 0.88 \pm 0.03 & 0.17 \pm 0.03 & 0.88 \pm 0.03 & 0.17 \pm 0.05 \\
        LSTM        & 0.64 \pm 0.01 & 0.00 \pm 0.00 & 0.61 \pm 0.05 & 0.42 \pm 0.24 & 0.73 \pm 0.18 & 0.17 \pm 0.15 & 0.84 \pm 0.12 & 0.42 \pm 0.15 \\
        Transformer & 0.85 \pm 0.12 & 0.00 \pm 0.00 & 0.44 \pm 0.38 & 0.36 \pm 0.37 & 0.83 \pm 0.11 & 0.75 \pm 0.11 & 0.87 \pm 0.09 & 0.00 \pm 0.00 \\
        ViT         & 0.93 \pm 0.09 & 0.00 \pm 0.00 & 0.84 \pm 0.11 & 0.01 \pm 0.02 & 0.89 \pm 0.04 & 0.92 \pm 0.08 & 0.82 \pm 0.03 & 0.25 \pm 0.06 \\
        \bottomrule
    \end{tabular}%
    } 
\end{table*}

\subsection{Task-Specific Metrics Analysis}

The primary goal was to develop models that perform well on metrics critical to the assembly process: precision for the mounted class and recall for the jammed class. A high mounted precision minimizes false positives – incorrectly predicting a successful assembly that actually failed (e.g., jammed). Such errors lead to undetected faulty assemblies, causing potential rework or safety issues. High jammed recall minimizes false negatives – failing to detect a jammed fastener. Reliable detection allows immediate process halting, preventing damage and delays. Therefore, evaluating model performance requires focusing on these specific metrics rather than overall accuracy, especially given the dataset imbalance. We also investigated whether including rotational angle data alongside force/torque data improved performance on these critical metrics.

Examining the results using only force/torque data, models trained on the original, imbalanced dataset often exhibited poor jammed recall (0.00±0.00 for MLP, LSTM, Transformer, ViT), highlighting the limitation of standard training for critical failure detection. While mounted precision was generally higher (e.g., ViT: 0.95±0.10), this could be misleading due to the failure to detect jams. Addressing imbalance via SMOTE (Balanced/Synthetic datasets) significantly improved jammed recall (e.g., Transformer on Synthetic: 0.75±0.09) while maintaining or improving mounted precision (e.g., CNN on Balanced: 0.90±0.03). Class weighting (CW) also boosted jammed recall (e.g., MLP: 0.81±0.30) but often severely reduced mounted precision (e.g., Transformer: 0.75 to 0.40) and increased variance.

The inclusion of rotational data yielded mixed and model-dependent results regarding these task-specific metrics. It did not universally improve performance. For mounted precision, adding rotation sometimes caused slight decreases (e.g., CNN original: 0.87 vs 0.82; ViT original: 0.95 vs 0.93), occasionally slight increases (e.g., CNN CW: 0.74 vs 0.83; ViT Balanced: 0.87 vs 0.89), and sometimes had minimal effect (e.g., MLP original/balanced/synthetic).
However, adding rotation occasionally provided a more significant benefit for jammed recall, particularly under certain data treatments. For instance, with class weighting, CNN's jammed recall improved from 0.20±0.21 to 0.48±0.29 when rotation was added. Most critically, the ViT model trained on the balanced dataset achieved its highest jammed recall (0.92±0.08) only when rotational data was included (compared to 0.33±0.10 without rotation). Similarly, the Transformer on the balanced dataset achieved a high jammed recall (0.75±0.11) with rotation.

Overall, while force/torque data appears to contain substantial predictive information, the added rotational data can be beneficial for improving jammed recall in specific model/data configurations (notably ViT-Balanced, CNN-CW, Transformer-Balanced), which is crucial for safety and damage prevention. However, this potential improvement must be weighed against the lack of consistent benefit for mounted precision and the significant increase in model complexity often associated with processing the additional data channel (e.g., ViT parameters increased substantially). Therefore, the choice of including rotational data should be based on carefully evaluating this trade-off for the specific application priorities.

\subsection{Model Complexity vs. Performance Trade-off}

The models evaluated represent a wide range of complexities, measured by the number of trainable parameters, shown in Table \ref{tab:number-params}. MLP and ViT models generally required the largest number of parameters, often scaling into the millions, especially ViT when trained on larger (balanced or synthetic) datasets with rotation (16.7M and 29.5M parameters, respectively). CNNs varied significantly; on the original dataset, they were complex (e.g., 6.5M parameters without rotation), but hyperparameter optimization on balanced or synthetic datasets yielded much lighter models (often under 100k parameters) while achieving top performance (e.g., 0.90±0.03 mounted precision with 94k parameters for CNN without rotation on the balanced dataset). LSTM and Transformer models consistently had the lowest parameter counts, often in the hundreds or tens of thousands, making them computationally efficient.

\begin{table}[ht] 
    \centering
    \caption{Number of Parameters (Millions M or Thousands k) per model, rotation condition, and data treatment.} 
    \label{tab:number-params} 
    \sisetup{table-align-uncertainty=true, separate-uncertainty=true} 
    \begin{tabular}{lccccc}
        \toprule
        \textbf{Model} & \textbf{Rotation} & {\textbf{Orig.}} & {\textbf{CW}} & {\textbf{Bal.}} & {\textbf{Syn.}} \\
        \midrule
        \multirow{2}{*}{MLP} & No & 1.41~M &  3.30~M &  7.31 M &  4.24 M \\
                             & Yes & 5.21 M &  3.57 M &  4.27 M &  8.22 M \\
        \midrule
        \multirow{2}{*}{CNN} & No & 6.55 M &  59.0 k &  94.4 k &  83.0 k \\
                             & Yes & 3.83 M &  75.2 k &  78.6 k & 292.8 k \\
        \midrule
        \multirow{2}{*}{LSTM} & No & 0.3 k &   1.3 k &   0.4 k &   1.3 k \\
                              & Yes & 0.3 k &   1.1 k &   0.7 k &   0.3 k \\
        \midrule
        \multirow{2}{*}{Transformer} & No & 4.0 k &   8.0 k &  10.3 k &  22.2 k \\
                                     & Yes & 38.2 k &   2.8 k &  42.1 k &  67.9 k \\
        \midrule
        \multirow{2}{*}{ViT} & No & 14.8 M &  7.04 M &  1.53 M &  1.12 M \\
                             & Yes & 7.81 M &  2.30 M & 16.7 M  & 29.5 M  \\
        \bottomrule
    \end{tabular}

\end{table}

There wasn't always a direct correlation between complexity and performance. While the highly parameterized ViT performed well on the original data (mounted precision 0.95±0.10 without rotation, 0.93±0.09 with rotation), its balanced/synthetic data performance wasn't consistently superior to the much lighter CNN or Transformer models. For instance, the Transformer without rotation achieved 0.86±0.08 mounted precision and a high 0.75±0.09 jammed recall on the synthetic dataset with only 22k parameters. The CNN without rotation achieved 0.90±0.03 mounted precision and 0.25±0.04 jammed recall on the balanced dataset with 94k parameters. This suggests that for this specific task, architectures like CNN and Transformer can capture the relevant temporal features effectively without excessive parameterization, offering a better trade-off between performance and computational cost compared to MLP or ViT in many scenarios. LSTM models, despite their low complexity, generally underperformed on mounted precision compared to other architectures.

\subsection{Data Imbalance Handling}

The raw dataset suffers from significant class imbalance (63.9\% Mounted, 23.4\% Not Mounted, 12.7\% Jammed). We compared four approaches: training on the original data, using class weighting (CW) on the original data, oversampling minority classes using SMOTE to create a balanced dataset, and further oversampling with SMOTE to create a larger synthetic dataset.

Training on the original data resulted in poor jammed recall across most models, as they were biased towards the majority mounted class. Applying class weighting (Original CW) yielded mixed results. It significantly improved jammed recall for some models (e.g., MLP without rotation: 0.81±0.30; MLP with rotation: 0.57±0.26; CNN with rotation: 0.48±0.29) compared to the original data, but often drastically reduced mounted precision (e.g., Transformer without rotation dropped from 0.75 to 0.40; Transformer with rotation dropped from 0.85 to 0.44) and introduced high variance (large standard deviations).

Oversampling with SMOTE (Balanced and Synthetic datasets) generally provided a more stable improvement. The balanced dataset often led to the best mounted precision (e.g., CNN without rotation: 0.90±0.03; CNN with rotation: 0.88±0.03; ViT with rotation: 0.89±0.04) while also boosting jammed recall compared to the original data, although not always as high as the CW approach (e.g., ViT with rotation on balanced: 0.92±0.08 jammed recall). The synthetic dataset sometimes further improved jammed recall (e.g., Transformer without rotation: 0.75±0.09) but didn't consistently improve mounted precision over the balanced dataset, suggesting diminishing returns from simply quadrupling the SMOTE-generated data. Notably, the Transformer model benefited significantly from the synthetic data regarding jammed recall. Overall, SMOTE-based oversampling appears more effective and stable than class weighting for this task, improving minority class detection without catastrophically impacting majority class precision.

\subsection{Model selection and statistical analysis}

Selecting the best model depends on the specific requirements of the assembly task, particularly the tolerance for different types of errors. Based on a balance between high mounted precision and high jammed recall, while considering model complexity, three candidates stood out: ViT with rotation on the Balanced dataset, Transformer without rotation on the Synthetic dataset, and CNN without rotation on the Balanced dataset.

To rigorously compare these contenders, paired t-tests were performed on the 10-fold cross-validation results. The results revealed no statistically significant differences between the three models regarding mounted precision. ViT vs. Transformer yielded p=0.2419, ViT vs. CNN yielded p=0.6022, and Transformer vs. CNN yielded p=0.0830. Although the difference between the Transformer and CNN approaches is significant, we cannot conclude a difference based on this metric alone. This indicates that all three top models perform similarly well in correctly identifying mounted cases among their positive predictions. Analyzing the jammed recall found highly statistically significant differences. ViT significantly outperformed both Transformer (p<0.001) and CNN (p<0.001). Furthermore, Transformer significantly outperformed CNN (p<0.001).

The statistical analysis strongly favors models with superior jammed recall, as this metric is directly linked to preventing critical failures. The ViT with rotation trained on the balanced dataset demonstrated statistically significantly higher jammed recall (0.92±0.08) than the other top contenders, making it the preferred choice if maximizing the detection of jammed fasteners is the highest priority, despite its high computational complexity (16.7M parameters). If computational resources are constrained, the Transformer without rotation trained on the synthetic dataset offers a compelling alternative. It achieves statistically significantly better jammed recall (0.75±0.09) than the CNN model, performs comparably to the other models on mounted precision (0.86±0.08), and does so with remarkably low complexity (22k parameters). While achieving high mounted precision, the CNN model is statistically the weakest in detecting critical jammed failures. Ultimately, the ViT model is recommended as the best model because its statistically superior ability to detect jammed fasteners directly addresses the critical need to prevent potentially costly or unsafe assembly failures in applications like aircraft manufacturing.

\section{Conclusions}\label{sec:conclusions}

This paper addressed the failure detection and isolation in automated aircraft fastener assembly, specifically focusing on threaded collars where failures like jamming can lead to significant costs and safety concerns. Recognizing the limitations imposed by small, imbalanced datasets common in such industrial applications, we investigated the efficacy of deep learning models trained directly on raw, multivariate time-series sensor data (force, torque, and optionally, rotation angle) without manual feature extraction. Our methodology emphasized a task-oriented approach, shifting the evaluation focus from generic accuracy to metrics directly relevant to the assembly process. We systematically compared strategies for handling data imbalance, including class weighting and SMOTE-based oversampling, integrated within a rigorous hyperparameter optimization and 10-fold cross-validation framework.

Key findings demonstrate that addressing data imbalance is essential for reliable failure detection. While training on the original data yielded poor jammed recall across most architectures, SMOTE oversampling proved effective and stable, significantly boosting jammed recall while maintaining high mounted precision. The balanced dataset often provided the best mounted precision, while further synthetic data generation sometimes enhanced recall further, particularly for the Transformer model. Class weighting improved jammed recall but often detrimentally affected mounted precision. The inclusion of rotational data did not universally improve performance. Still, it was beneficial for jammed recall in specific configurations, notably enabling the ViT model on the balanced dataset to achieve the highest recall for this critical failure mode (0.92±0.08). Furthermore, our analysis revealed significant performance-complexity trade-offs, with lightweight models like the Transformer demonstrating competitive performance, especially when trained on augmented data (0.75±0.09 jammed recall with only 22k parameters).

Statistical analysis using paired t-tests confirmed that while the top-performing models showed no significant difference in mounted precision, there were highly significant differences (p<0.001) in their ability to detect jams. The ViT model with rotation on the balanced dataset significantly outperformed all others in jammed recall. Based on these statistically validated results, we recommend the ViT model with rotation trained on the balanced dataset as the most effective solution when maximizing the detection of critical jammed failures is the priority, despite its higher computational complexity. For applications with constrained computational resources, the Transformer model without rotation trained on the synthetic dataset offers an excellent, highly efficient alternative, providing significantly better jam detection than the CNN model while maintaining comparable precision.

This work demonstrates the viability of applying deep learning directly to raw sensor data for complex assembly FDI, highlighting the necessity of task-specific metric optimization and effective data augmentation strategies like SMOTE for handling imbalance. Future research should explore more advanced synthetic data generation techniques like generative methods tailored for temporal dependencies, investigate anomaly detection methods to identify unforeseen failure modes, and apply interpretable machine learning techniques to understand model decision-making. This could potentially inform better tool design and control strategies to prevent failures proactively.


\section*{Acknowledgment}
The authors thank CNPQ grant 141395/2017-6, the Coordenação de Aperfeiçoamento de Pessoal de Nível Superior—Brazil (CAPES) -- Finance Code 001, and the EMBRAER-GPX for their contributions to the dataset.

\bibliographystyle{unsrt}  
\bibliography{references}

\begin{thebibliography}{10}

\bibitem{Gonzaga2019}
Andr{\'{e}}~V.S. Silva and Lu{\'{i}}s~Gonzaga Trabasso.
\newblock {Design for Automation within the aeronautical domain}, jul 2019.

\bibitem{Airbus2019assembly}
Airbus.
\newblock {Airbus inaugurates new A320 structure assembly line in Hamburg}, 2019.

\bibitem{Gates2019}
Dominic Gates.
\newblock {Boeing abandons its failed fuselage robots on the 777X, handing the job back to machinists}, 2019.

\bibitem{Kheddar2019humanoidsaircraft}
Abderrahmane Kheddar, Maximo~A. Roa, Pierre~Brice Wieber, Francois Chaumette, Fabien Spindler, Giuseppe Oriolo, Leonardo Lanari, Adrien Escande, Kevin Chappellet, Fumio Kanehiro, Patrice Rabate, Stephane Caron, Pierre Gergondet, Andrew Comport, Arnaud Tanguy, Christian Ott, Bernd Henze, George Mesesan, and Johannes Englsberger.
\newblock {Humanoid Robots in Aircraft Manufacturing: The Airbus Use Cases}.
\newblock {\em IEEE Robotics and Automation Magazine}, 26(4):30--45, dec 2019.

\bibitem{Boeing2013nparts}
Boeing.
\newblock {Boeing Celebrates Delivery of 50th 747-8}, 2013.

\bibitem{Lisi-aerospace2021}
Lisi-aerospace.
\newblock {HI-LOK™ and HI-LITE™ Nuts and Collars}, 2021.

\bibitem{Kuka2021screw1}
Kuka.
\newblock {KUKA cell4\_screwsetting}, 2021.

\bibitem{Kuka2021screw2}
Kuka.
\newblock {KUKA ready2\_fasten\_micro}, 2021.

\bibitem{STP-CONCEPT2021screw3}
STP-CONCEPT.
\newblock {SCREWING ROBOT}, 2021.

\bibitem{Duepi2021screw4}
Duepi.
\newblock {Screwing robots}, 2021.

\bibitem{Denso2021screw5}
Denso.
\newblock {Case Studies Robot Application Screwing}, 2021.

\bibitem{Drouot2018}
Adrien Drouot, Ran Zhao, Lucas Irving, David Sanderson, and Svetan Ratchev.
\newblock Measurement assisted assembly for high accuracy aerospace manufacturing.
\newblock {\em IFAC-PapersOnLine}, 51:393--398, 2018.

\bibitem{Mosqueira2012}
G.~Mosqueira, J.~Apetz, K.M. Santos, E.~Villani, R.~Suterio, and L.G. Trabasso.
\newblock Analysis of the indoor gps system as feedback for the robotic alignment of fuselages using laser radar measurements as comparison.
\newblock {\em Robotics and Computer-Integrated Manufacturing}, 28:700--709, 12 2012.

\bibitem{FAA2012}
FAA.
\newblock {\em Aviation Maintenance Technician Handbook-Airframe Volume 1}.
\newblock Federal Aviation Administration, 2012.

\bibitem{Hardy2019}
David~F. Hardy and David~L. DuQuesnay.
\newblock Effect of repetitive collar replacement on the residual strength and fatigue life of retained hi-lok fastener pins.
\newblock {\em Metals}, 9:445, 4 2019.

\bibitem{Mason2018}
Zhenzhong Jia, Ankit Bhatia, Reuben~M. Aronson, David Bourne, and Matthew~T. Mason.
\newblock {A Survey of Automated Threaded Fastening}.
\newblock {\em IEEE Transactions on Automation Science and Engineering}, pages 1--13, 2018.

\bibitem{Mason2010forcesignature}
Alberto Rodriguez, David Bourne, Mathew Mason, Gregory~F. Rossano, and {JianJun Wang}.
\newblock {Failure detection in assembly: Force signature analysis}.
\newblock In {\em 2010 IEEE International Conference on Automation Science and Engineering}, pages 210--215. IEEE, aug 2010.

\bibitem{McIntyre2005}
M.L. McIntyre, W.E. Dixon, D.M. Dawson, and I.D. Walker.
\newblock {Fault identification for robot manipulators}.
\newblock {\em IEEE Transactions on Robotics}, 21(5):1028--1034, oct 2005.

\bibitem{Arrichiello2015}
Filippo Arrichiello, Alessandro Marino, and Francesco Pierri.
\newblock {Observer-Based Decentralized Fault Detection and Isolation Strategy for Networked Multirobot Systems}.
\newblock {\em IEEE Transactions on Control Systems Technology}, 23(4):1465--1476, jul 2015.

\bibitem{Diryag2014}
Ali Diryag, Marko Miti{\'{c}}, and Zoran Miljkovi{\'{c}}.
\newblock {Neural networks for prediction of robot failures}.
\newblock {\em Proceedings of the Institution of Mechanical Engineers, Part C: Journal of Mechanical Engineering Science}, 228(8):1444--1458, jun 2014.

\bibitem{Van2013}
Mien Van, Hee~Jun Kang, Young~Soo Suh, and Kyoo~Sik Shin.
\newblock {A robust fault diagnosis and accommodation scheme for robot manipulators}.
\newblock {\em International Journal of Control, Automation and Systems}, 11(2):377--388, 2013.

\bibitem{Ferhat2021}
Mahmoud Ferhat, Mathieu Ritou, Philippe Leray, and Nicolas {Le Du}.
\newblock {Incremental discovery of new defects: application to screwing process monitoring}.
\newblock {\em CIRP Annals}, jun 2021.

\bibitem{Moreira2018}
Guilherme~R. Moreira, Gustavo J.~G. Lahr, Thiago Boaventura, Jose~O. Savazzi, and Glauco A.~P. Caurin.
\newblock {Online prediction of threading task failure using Convolutional Neural Networks}.
\newblock In {\em 2018 IEEE/RSJ International Conference on Intelligent Robots and Systems (IROS)}, pages 2056--2061, Madrid, oct 2018. IEEE.

\bibitem{Cheng2019}
Xianyi Cheng, Zhenzhong Jia, and Matthew~T. Mason.
\newblock {Data-Efficient Process Monitoring and Failure Detection for Robust Robotic Screwdriving}.
\newblock In {\em 2019 IEEE 15th International Conference on Automation Science and Engineering (CASE)}, pages 1705--1711. IEEE, aug 2019.

\bibitem{Aronson2017}
Reuben~M. Aronson, Ankit Bhatia, Zhenzhong Jia, Mathieu Guillame-Bert, David Bourne, Artur Dubrawski, and Matthew~T. Mason.
\newblock {Data-Driven Classification of Screwdriving Operations}.
\newblock In {\em 2016 International Symposium on Experimental Robotics. ISER 2016. Springer Proceedings in Advanced Robotics}, pages 244--253. Springer, Cham, 2017.

\bibitem{Cheng2018}
Xianyi Cheng, Zhenzhong Jia, Ankit Bhatia, Reuben~M. Aronson, and Matthew~T. Mason.
\newblock {Sensor Selection and Stage {\&} Result Classifications for Automated Miniature Screwdriving}.
\newblock In {\em 2018 IEEE/RSJ International Conference on Intelligent Robots and Systems (IROS)}, pages 6078--6085. IEEE, oct 2018.

\bibitem{Barella2021andrePonce}
Victor~H. Barella, Luís~P.F. Garcia, Marcilio~C.P. de~Souto, Ana~C. Lorena, and André~C.P.L.F. de~Carvalho.
\newblock Assessing the data complexity of imbalanced datasets.
\newblock {\em Information Sciences}, 553:83--109, 4 2021.

\bibitem{dosovitskiy2021image}
Alexey Dosovitskiy, Lucas Beyer, Alexander Kolesnikov, Dirk Weissenborn, Xiaohua Zhai, Thomas Unterthiner, Mostafa Dehghani, Matthias Minderer, Georg Heigold, Sylvain Gelly, Jakob Uszkoreit, and Neil Houlsby.
\newblock An image is worth 16x16 words: Transformers for image recognition at scale.
\newblock {\em arXiv preprint arXiv:2010.11929}, 2021.

\bibitem{Godoy2022}
Ricardo~V. Godoy, Gustavo J.~G. Lahr, Anany Dwivedi, Tharik J.~S. Reis, Paulo~H. Polegato, Marcelo Becker, Glauco A.~P. Caurin, and Minas Liarokapis.
\newblock Electromyography-based, robust hand motion classification employing temporal multi-channel vision transformers.
\newblock {\em IEEE Robotics and Automation Letters}, 7:10200--10207, 10 2022.

\bibitem{Seneviratne2001}
L~D Seneviratne, F~A Ngemoh, S~W~E Earles, and K~A Althoefer.
\newblock {Theoretical modelling of the self-tapping screw fastening process}.
\newblock {\em Proceedings of the Institution of Mechanical Engineers, Part C: Journal of Mechanical Engineering Science}, 215(2):135--154, feb 2001.

\bibitem{Wiedmann2006spatialKinAnalysis}
Stephen Wiedmann and Bob Sturges.
\newblock {Spatial Kinematic Analysis of Threaded Fastener Assembly}.
\newblock {\em Journal of Mechanical Design}, 128(1):116, 2006.

\bibitem{Wiedmann2006threadStarting}
Stephen Wiedmann and Bob Sturges.
\newblock {A Full Kinematic Model of Thread-Starting for Assembly Automation Analysis}.
\newblock {\em Journal of Mechanical Design}, 128(1):128, 2006.

\bibitem{Nicolson1991}
E.J. Nicolson and R.S. Fearing.
\newblock {Dynamic modeling of a part mating problem: threaded fastener insertion}.
\newblock {\em Proceedings IROS '91:IEEE/RSJ International Workshop on Intelligent Robots and Systems '91}, pages 30--37, 1991.

\bibitem{Nicolson1993}
E.J. Nicolson and R.S. Fearing.
\newblock {Compliant control of threaded fastener insertion}.
\newblock {\em [1993] Proceedings IEEE International Conference on Robotics and Automation}, pages 484--490, 1993.

\bibitem{Lara2000}
B.~Lara, L.D. Seneviratne, and K.~Althoefer.
\newblock {Radial basis artificial neural networks for screw insertions classification}.
\newblock In {\em Proceedings 2000 ICRA. Millennium Conference. IEEE International Conference on Robotics and Automation. Symposia Proceedings (Cat. No.00CH37065)}, volume~2, pages 1912--1917. IEEE, 2000.

\bibitem{Althoefer2005}
Kaspar Althoefer, Bruno Lara, and Lakmal~D. Seneviratne.
\newblock {Monitoring of Self-Tapping Screw Fastenings Using Artificial Neural Networks}.
\newblock {\em Journal of Manufacturing Science and Engineering}, 127(1):236--243, feb 2005.

\bibitem{Ruusunen2003}
Mika Ruusunen and Marko Paavola.
\newblock {Monitoring of Automated Screw Insertion Processes-A Soft Computing Approach}.
\newblock {\em IFAC Proceedings Volumes}, 36(3):169--174, apr 2003.

\bibitem{Althoefer2008}
K.~Althoefer, B.~Lara, Y~H Zweiri, and L.~D. Seneviratne.
\newblock {Automated failure classification for assembly with self-tapping threaded fastenings using artificial neural networks}.
\newblock {\em Proceedings of the Institution of Mechanical Engineers, Part C: Journal of Mechanical Engineering Science}, 222(6):1081--1095, jun 2008.

\bibitem{Matsuno2013}
Takayuki Matsuno, Jian Huang, and Toshio Fukuda.
\newblock {Fault detection algorithm for external thread fastening by robotic manipulator using linear support vector machine classifier}.
\newblock In {\em 2013 IEEE International Conference on Robotics and Automation}, pages 3443--3450. IEEE, may 2013.

\bibitem{Teixeira2022}
Humberto~Nuno Teixeira, Isabel Lopes, Ana~Cristina Braga, Pedro Delgado, and Cristina Martins.
\newblock {\em {Screwing Process Monitoring Using MSPC in Large Scale Smart Manufacturing}}.
\newblock Springer, Cham., 2022.

\bibitem{Pastor2021threadQualityImbalanced}
Jose~F. Diez-Pastor, Alain Gil~Del Val, Fernando Veiga, and Andres Bustillo.
\newblock High-accuracy classification of thread quality in tapping processes with ensembles of classifiers for imbalanced learning.
\newblock {\em Measurement}, 168:108328, 1 2021.

\bibitem{Lahr2023dataset}
Gustavo Jose~Giardini Lahr, Thiago Henrique~Segreto Silva, Guilherme~Ribeiro Moreira, Thiago Boaventura, Glauco~Augusto de~Paula~Caurin, and Arash Ajoudani.
\newblock Kinematic and dynamic data from a robotic assembly of aeronautical threaded fasteners.
\newblock {\em Data in Brief}, 51:109674, 2023.

\bibitem{Mortaz2020}
Ebrahim Mortaz.
\newblock Imbalance accuracy metric for model selection in multi-class imbalance classification problems.
\newblock {\em Knowledge-Based Systems}, 210:106490, 12 2020.

\bibitem{Vaswani2017}
Ashish Vaswani, Noam Shazeer, Niki Parmar, Jakob Uszkoreit, Llion Jones, Aidan~N. Gomez, Lukasz Kaiser, and Illia Polosukhin.
\newblock {Attention Is All You Need}.
\newblock {\em Advances in Neural Information Processing Systems}, jun 2017.

\bibitem{optuna_2019}
Takuya Akiba, Shotaro Sano, Toshihiko Yanase, Takeru Ohta, and Masanori Koyama.
\newblock Optuna: A next-generation hyperparameter optimization framework.
\newblock In {\em Proceedings of the 25rd {ACM} {SIGKDD} International Conference on Knowledge Discovery and Data Mining}, page 2623–2631, 2019.

\bibitem{ChaoHong2017paa}
ChaoHong Ma, XiaoQing Weng, and ZhongNan Shan.
\newblock Early classification of multivariate time series based on piecewise aggregate approximation.
\newblock In Siuly Siuly, Zhisheng Huang, Uwe Aickelin, Rui Zhou, Hua Wang, Yanchun Zhang, and Stanislav Klimenko, editors, {\em Health Information Science}, pages 81--88, Cham, 2017. Springer International Publishing.

\bibitem{Shorten2019dataAugComputerVision}
Connor Shorten and Taghi~M. Khoshgoftaar.
\newblock A survey on image data augmentation for deep learning.
\newblock {\em Journal of Big Data}, 6:60, 12 2019.

\bibitem{Liu2020dataAugNLP}
Pei Liu, Xuemin Wang, Chao Xiang, and Weiye Meng.
\newblock A survey of text data augmentation.
\newblock In {\em 2020 International Conference on Computer Communication and Network Security (CCNS)}, pages 191--195. IEEE, 8 2020.

\bibitem{Chawla2002smote}
N.~V. Chawla, K.~W. Bowyer, L.~O. Hall, and W.~P. Kegelmeyer.
\newblock Smote: Synthetic minority over-sampling technique.
\newblock {\em Journal of Artificial Intelligence Research}, 16:321--357, 6 2002.

\bibitem{johnson2019survey}
Justin~M Johnson and Taghi~M Khoshgoftaar.
\newblock Survey on deep learning with class imbalance.
\newblock {\em Journal of Big Data}, 6(1):1--54, 2019.

\bibitem{ling2008cost}
Charles~X Ling and Victor~S Sheng.
\newblock Cost-sensitive learning and the class imbalance problem.
\newblock {\em Encyclopedia of machine learning}, 2011:231--235, 2008.

\end{thebibliography}

\end{document}